\documentclass[10pt, a4paper, oneside]{article}
\usepackage[hidelinks]{hyperref}
\usepackage{graphicx}
\usepackage{url}
\usepackage{ulem}
\usepackage{mathtools}
\usepackage{scalerel}
\usepackage{setspace}
\usepackage[strict]{changepage}
\usepackage{caption}
\usepackage[letterspace=-50]{microtype}
\usepackage{afterpage}
\usepackage{ragged2e}
\usepackage{algorithm}
\usepackage{algorithmic}
\usepackage{makecell}
\usepackage{multirow}
\usepackage[T1]{fontenc}
\usepackage{lmodern}
\usepackage{authblk}

\usepackage{titlesec}

\titleformat*{\section}{\Large\bfseries}
\titleformat*{\subsection}{\normalsize\bfseries}

\graphicspath{{./figures/}}

\usepackage[textwidth=8cm, margin=0cm, left=4.6cm, right=4.2cm, top=3.9cm, bottom=6.8cm, a4paper, headheight=0.5cm, headsep=0.5cm]{geometry}
\usepackage{fancyhdr}
\usepackage[format=plain, labelfont=it, textfont=it, justification=centering]{caption}

\apptocmd{\frame}{}{\justifying}{}

\newcommand\paperauthor{{Kozal, J., Guzy, F., Woźniak, M.: }}

\newcommand\papertitle{Employing chunk size adaptation to overcome concept drift}

\begin{document}

\title{{\fontsize{14pt}{14pt}\selectfont{\vspace*{-3mm}\papertitle\vspace*{-1mm}}}}


\author[1]{Jędrzej Kozal}
\author[1]{Filip Guzy}
\author[1]{Michał Woźniak}
\affil[1]{Wroc\l{}aw University of Science and Technology}
\affil[ ]{\textit {\{jedrzej.kozal,filip.guzy,michal.wozniak\}@pwr.edu.pl}}

\date{}

\label{first}
\maketitle

{\fontfamily{ptm}\selectfont
\begin{abstract}
{\fontsize{9pt}{9pt}\selectfont{\vspace*{-2mm}

Modern analytical systems must be ready to process streaming data and correctly respond to data distribution changes. The phenomenon of changes in data distributions is called \textit{concept drift}, and it may harm the quality of the used models. Additionally, the possibility of concept drift appearance causes that the used algorithms must be ready for the continuous adaptation of the model to the changing data distributions.
This work focuses on non-stationary data stream classification, where a classifier ensemble is used. To keep the ensemble model up to date, the new base classifiers are trained on the incoming data blocks and added to the ensemble while, at the same time, outdated models are removed from the ensemble. One of the problems with this type of model is the fast reaction to changes in data distributions. We propose a new \textit{Chunk Adaptive Restoration} framework that can be adapted to any block-based data stream classification algorithm. The proposed algorithm adjusts the data chunk size in the case of concept drift detection to minimize the impact of the change on the predictive performance of the used model.
The conducted experimental research, backed up with the statistical tests, has proven that \textit{Chunk Adaptive Restoration} significantly reduces the model's restoration time.}}
\end{abstract}}




\section{Introduction}

\textit{Data stream mining} focuses on the knowledge extraction from streaming data, mainly for the predictive model construction aimed at assigning arriving instances to one of the predefined categories. This process is characterized by additional difficulties that arise when data distribution evolves over time.
It is visible in many practical tasks as spam detection, where the spammers still change the message format to cheat anti-spam systems. Another example is medical diagnostics, where new SARS-CoV-2 mutations may cause different symptoms, which forces doctors to adapt and improve diagnostic methods \cite{Harvey2021}.

The mentioned above phenomenon is called \textit{concept drift}, and its nature can vary due to both the character and the rapidity. It forces classification models to adapt to new data characteristics and forget old, useless concepts.
An important characteristic of such systems is their reaction to the concept drift phenomenon, i.e., how much predictive performance deteriorates when it occurs and when the classification system will obtain the approved predictive quality for the new concept.
We should also consider another limitation: the classification system should be ready to classify incoming objects immediately, and dedicated computing and memory resources are limited.

Data processing models used by stream data classification systems can be roughly divided into two categories: online (object by object) processing (online learners), or block-based (chunk by chunk) data processing (block-based learners) \cite{Krawczyk:2017}. Online learners require model parameters to be updated when a new object appears, while the block-based method requires updates once per batch. The advantage of online learners is their fast adaptation to concept drift. However, in many practical applications, the effort of necessary computation (related to updating models after processing each object) is unacceptable. The model update can require many operations that involve changing data statistics, updating the model's internal structure, or learning a new model from scratch. These requirements can become prohibitive for high-velocity streams. 
Hence, more popular is block-based data processing, which requires less computational effort. However, it limits the model's potential for quick adaptation to changes in data distribution and fast restoration of performance after concept drift. 
In consequence, a significant problem is the proper selection of the chunk size. Smaller data block size results in faster adaptation. However, it increases the overall computing load. On the other hand, larger data chunks require less computation but result in a lower adaptive capacity of the classification model. Another valid consideration is the impact of chunk size on prediction stability. Models trained on smaller chunks typically have larger prediction variance, while models trained with larger chunks tend to have more stable predictions when the data stream is stationary. If concept drift occurs, a larger chunk increase probability that the data from different concepts will be placed in the same batch. 
Hence, selecting the chunk size is a trade-off encompassing computation power, adaptation speed, and predictions variance.

The trade-off described above includes features that are equally desired in many applications. Especially consumption of computation power and adaptation speed are both important when processing large data streams. We propose a new method that alleviates the downfalls of choosing between small or large chunk sizes by dynamically changing the current batch size. More precisely, our work introduces the \textit{Chunk-Adaptive Restoration} (CAR), a framework based on combined drift and stabilization detection techniques that adjusts the chunk sizes during the \textit{concept drift}. This approach slightly redefines the previous problem based on the observation that for many practical classification tasks, a period of changes in data distributions is followed by stabilization. Hence, we propose that when the \textit{concept drift} occurs, the model should be quickly upgraded, i.e., the data should be processed in small chunks, and during the stabilization period, the data block size may be extended. The advantage of the proposed method is its universality and the possibility of using it with various chunk-based data stream classifiers.

This work offers the following contributions:
\begin{itemize}
    \item Proposing the \textit{Chunk-Adaptive Restoration} framework to empower fluent restoration after \textit{concept drift} appearance.
    \item Formulating the \textit{Variance-based Stabilization Detection Method}, a technique complementary to all \textit{concept drift} detectors that simplifies chunk size adaptation and metrics calculation.
    \item Employing \textit{Chunk-Adaptive Restoration} for the adaptive data chunk size setting for selected state-of-the-art algorithms.
    \item Introducing a new stream evaluation metric, \textit{Sample Restoration}, to show the gains of the proposed methods.
    \item Experimental evaluation of the proposed approach based on various synthetic and real data streams and a detailed evaluation of its usefulness for the selected state-of-art methods.
\end{itemize}

\section{Related works}

This section provides a review of the related works. Firstly, we will discuss challenges specific to the learning from non-stationary data streams. Next, we discuss different methods of processing data streams. Following, we describe existing drift detection algorithms and ensemble methods. We continue by reviewing existing evaluation protocols and computational and memory requirements. We conclude this section by providing examples of other data stream learning methods that employ variable chunk size.

\subsection{Challenges related to data stream mining}

A data stream is a sequence of objects described by their attributes. In the case of a classification task, each learning object should be labeled. The number of items may be vast, potentially infinite. Observations in the stream may arrive at different times, and the time intervals between their arrival could vary considerably. The main differences between analyzing data streams and static datasets include \cite{Bifet:2018}:
\begin{itemize}
\item No one can control the order of incoming objects
\item The computation resources are limited, but the analyzer should be ready to process the incoming item in a~reasonable time
\item The memory resources are also limited, but the data stream size may be huge or infinite, which causes memorizing all the items impossible 
\item Data streams are susceptible to change, i.e., data distributions may change over time
\item The labels of arriving items are not for free, for some cases impossible to get, or available with delay (e.g., in banking for credit approval task after a few years)
\end{itemize}

The canonical classifiers usually do not consider that the probabilistic characteristics of the classification task may evolve \cite{Duda:2001}. Such a phenomenon is known as \textit{concept drift} \cite{Widmer:1996} and a few concept drift taxonomies have been proposed. The most popular consider how rapid the drift is, then we can distinguish sudden drift and incremental one. An additional difficulty is a case when, during the transition between two concepts, objects from two different concepts appear for some time simultaneously (gradual drift). 
We can also take into consideration the influence of the probabilistic characteristics on the classification task \cite{Gama:2014}:
\begin{itemize}
\item virtual concept drift does not impact the decision boundaries but affects the probability density functions \cite{Widmer:1993}, and  Widmer and Kubat \cite{Widmer:1996} imputed it rather to incomplete data representation than to the true changes in concepts, 
\item real concept drift affects the \textit{posterior }probabilities and may impact the unconditional probability density function \cite{Widmer:1996}.
\end{itemize}



\subsection{Methods for processing data streams}

The data stream can be divided into small portions of the data called data chunks. This method is known as batch-based or chunk-based learning. Choosing the proper size of the chunk is crucial because it may significantly affect the classification \cite{Junsawang:2019}. Unfortunately, the unpredictable appearance of the concept drift makes it difficult. Several approaches may help overcome this problem, e.g., using different windows for processing data \cite{Lazarescu:2004} or adjusting chunk size dynamically \cite{Widmer:1996}. Unfortunately, most chunk-based classification methods assume that the size of the data chunk is priorly set and remains unchanged during the data processing.

Instead of chunk-based learning, the algorithm can learn incrementally (online) as well. Training examples arrive one by one at a given time, and they are not kept in memory. The advantage of this solution is the need for small memory resources. However, the effort of necessary computation related to updating models after processing each individual object is unacceptable, especially in the high-velocity data streams, i.e., Internet of Things (IoT) applications.

When processing a non-stationary data stream, we can rely on a drift detector to point moments when data distribution has changed and take appropriate actions. The alternative is to use inherent adaptation properties of models (update \& forget). In the following subsection, we will discuss both of these approaches.

\subsection{Drift detection methods}

A drift detector is an algorithm that can inform about any changes taking place within data stream distributions. The data labels or a classifier's performance (measured using any metric, such as accuracy) is required to detect a real concept drift \cite{Sobolewski:2013}.  We have to realize that drift detection is a non-trivial task. The detection should be done as quickly as possible to replace an outdated model and minimize restoration time. On the other hand, false alarms are unacceptable, as they will lead to an incorrect model adaptation and resource spending where there is no need for it \cite{Gustafsson:2000}. \textit{\textsc{DDM} (Drift Detection Method)} \cite{Gama:2004} is one of the most popular detectors that incrementally estimates an error of a classifier. Because we assume the classifier training method's convergence, the error should decrease with the appearance of subsequent learning objects \cite{Raudys:2014}. If the reverse behavior is observed, then we may suspect a change of probability distributions. \textsc{DDM} uses the \textit{three-sigma rule} to detect a drift. \textit{{\textsc{EDDM}} (Early Drift Detection Methods)} \cite{Baena:2006} is an extension of \textsc{DDM}, where the window size selection procedure is based on the same heuristics. Additionally, the \textit{distance error rate} is being used instead of the classifier's error rate. Blanco et al. \cite{Blanco:2015} proposed very interesting drift detectors that use the non-parametric estimation of classifier error employing Hoeffding's and McDiarmid's inequalities.

\subsection{Ensemble methods}

One of the most promising data stream classification research directions, which usually employs chunk-based data processing is the classifier ensemble approach \cite{Krawczyk:2017}. Its advantage is that the classifier ensemble can easily adapt to the concept drift using different updating strategies \cite{Kuncheva:2004a}:
\begin{itemize}
\item \textit{Dynamic combiners} -- individual classifiers are trained in advance, and they are not updated anymore. The ensemble classifier adapts to changing data distribution by changing the combination rule parameters.
\item \textit{Updating training data} -- incoming examples are used to retrain  component classifiers (e.g., online bagging \cite{Oza:2008}).
\item \textit{Updating ensemble members} \cite{Bifet:2009,Rodriguez:2008}.
\item \textit{Changing ensemble lineup} -- replacing outdated classifiers in the ensemble, e.g., new individual models are trained on the most recent data and added to the ensemble. The ensemble pruning procedure is applied, which chooses the most valuable set of individual classifiers \cite{Jackowski:2014}.
\end{itemize}

A comprehensive overview of techniques using classifier ensemble \cite{Krawczyk:2017} was presented by Krawczyk et al. Let us shortly characterize some popular strategies used during the experiments.
\textit{Streaming Ensemble Algorithm} (\textsc{SEA}) \cite{Street:2001} is the simple classifier ensemble with changing lineup, where the individual classifiers are trained on the successive data chunks. To keep the model up-to-date, the base classifiers with the lowest accuracy are removed from the ensemble. 
Wang et al. proposed \textit{Accuracy Weighted Ensembles} (\textsc{AWE}) \cite{Wozniak:2013} employing the weighted voting rules, where weights depend on the accuracy obtained on the testing data. Brzezinski and Stefanowski proposed \textit{Accuracy Updated Ensemble} (\textsc{AUE}), which extends AWE by using online classifiers and updating them according to the current distribution \cite{Brzezinski:2011}. Wozniak et al. developed \textit{Weighted Aging Ensemble} (\textsc{WAE}), which trains base classifiers on successive data chunks, and the final decision is made on weighted voting, where weights depend on accuracy and ensemble diversity. This algorithm additionally employs the decoy function to decrease the weights of outdated individuals \cite{Wozniak:2013}.

\subsection{Existing evaluation methodology}

Because this work mainly focuses on improving classifier behavior after the concept drift appearance, apart from the classifier's predictive performance, we should also consider memory consumption, the time required to update the model, and time to decide. However, it should also be possible to evaluate how the model reacts to changes in the data distribution.  Shaker and H{\"{u}}llermeier \cite{Shaker:2015} presented a complete framework for evaluating the recovery rate, including the proposition of two metrics \textit{restoration time} and \textit{maximum performance loss}. In this framework, the notion of pure streams was introduced i.e., streams containing only one concept. Two pure streams $S_A$ and $S_B$ are mixed into third stream $S_C$, starting with concepts only from the first stream and gradually increasing a percentage of concepts from the second stream.
\textit{Restoration time} was defined as a length of the time interval between two events - first a performance measured on $S_C$ drops below 95\% of a $S_A$ performance, and then the performance on $S_C$ rise above 95\% of $S_B$ performance. The \textit{Maximum performance loss} is the maximum difference between $S_C$ performance and lowest performance on either $S_A$ or $S_B$.
Zliobaite et al. \cite{Zliobaite:2015} proposed that evaluating the profit from the model update should consider the memory and computing resources involved in its update.

\subsection{Computational and memory requirements}

While designing a data stream classifier, we should also consider the computation power and memory limitations and that we usually have limited access to data labels. These data stream characteristics pose the need for other algorithms than ones previously developed for \textit{batch learning}, where data are stored infinitely and persistently. Such learning algorithms cannot fulfill all data stream requirements, such as memory usage constraints, limited processing time, and one scan of incoming examples. However, simple incremental learning is usually insufficient, as it does not meet tight computational demands and does not tackle evolving nature of data sources \cite{Krempl:2014}.

Constraints on memory and time have resulted in different windowing techniques, sampling (e.g., reservoir sampling), and other summarization approaches. Also, we have to realize that when the concept drift appears, data from the past may become irrelevant or even harmful for the current models, deteriorating the predictive performance of the classifiers. Thus an appropriate implementation of a forgetting mechanism (where old data instances are discarded) is crucial.

\subsection{Other approaches that modify chunk size}

Dynamic chunk size adaptation was proposed in some works earlier \cite{wedcs} \cite{8924892} \cite{ADWIN}.
Liu et al. \cite{wedcs} utilize information about the occurrence of drift from drift detector. If drift occurs in the middle of the chunk, data is divided into two chunks, hence dynamic chunk size. If there is no drift inside the chunk, the whole batch is used. In the prepared chunk, the majority class is undersampled. A new classifier is trained and added to the ensemble, and older classifiers are updated. 
Lu et al. \cite{8924892} also utilize an ensemble framework for imbalanced stream learning. In this approach, chunk size grows incrementally. Two chunks are compared based on ensembles predictions variance. An algorithm for calculating prediction variance called \textit{subunderbagging} is introduced. Computed variance is compared using F-test. Chunk size increases if the p-value is less than a predefined threshold; otherwise, the whole ensemble is updated with the selected chunk size. The whole process repeats as long as the p-value is lower than the threshold.
In both of these works, dynamic chunk size was used as means of handling imbalanced data streams. In contrast, we show that changing chunk size can be beneficial when handling concept drifts in general. Therefore, we do not focus primarily on imbalanced data. 

Bifet et al. \cite{ADWIN} introduced a method for handling concept drift with varying chunk sizes. Each incoming chunk is divided into two parts: older and new. Empirical means of data in each subchunk are compared using Hoeffding bound. If the difference between two means exceeds the threshold defined by confidence value, then data in the older window is qualified as out of date and is dropped. Later window with data for current concept grows, until next drift is detected and data is split again. This approach allows for detecting drift inside the chunk.

\section{Methods}
\begin{figure}[h]
    \centering
    \includegraphics[width=\textwidth]{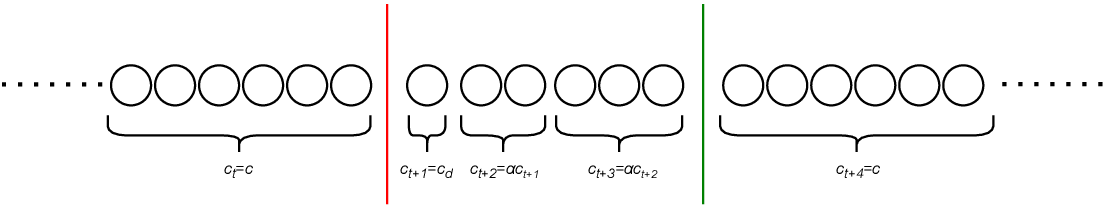}
    \captionsetup{font=small}
    \caption{\fontsize{10pt}{11pt}\selectfont{\itshape{Chunk-Adaptive Restoration visualization. Red line marks the concept drift, green line marks the stabilization.}}}
    \label{fig:carvis}
\end{figure}

This paper presents a general framework that can be used for training any chunk-based classifier ensemble. This approach aims to reduce the restoration time, i.e., a period needed to stabilize the classification model performance after concept drift occurs. As we mentioned, most methods assume a fixed data chunk size, which is a parameter of these algorithms. Our proposal does not modify the core of a learning algorithm itself. Still, based on the predictive performance estimated on a given data chunk, it only indicates what data chunk size is to be taken by a given algorithm in the next step. 
We provide schema of our method in Fig.~\ref{fig:carvis}.
The intuition tells us that after the occurrence of the concept drift, the size of the chunk should be small to quickly train new models that will replace the models learned on the data from the previous concept in the ensemble. When the stabilization is reached, the ensemble contains base models trained on data from a new concept. In this moment we can extend the chunk size so classifiers in the ensemble can achieve better performance and even greater stability by learning on larger portions of data from the streams because the analyzed concept is already stable. 

Let us present the proposed framework in detail.

\subsection{Chunk-Adaptive Restoration}

\begin{figure}[h]
    \centering
    \includegraphics[width=0.7\textwidth]{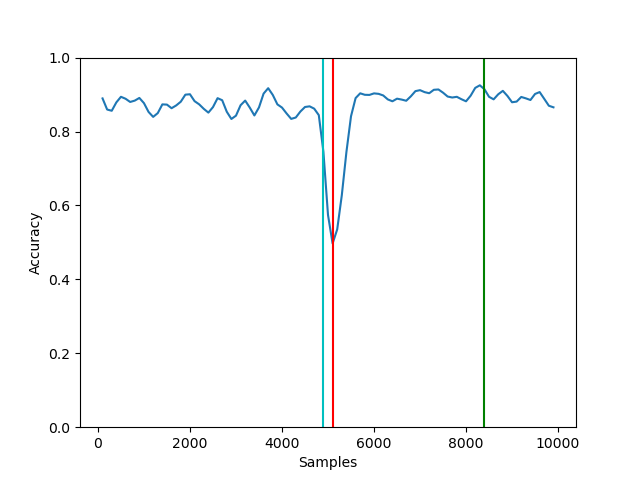}
    \captionsetup{font=small}
    \caption{\fontsize{10pt}{11pt}\selectfont{\itshape{Exemplary accuracy for data stream with abrupt concept. Red line denotes drift detection, green stabilization detection, and blue beginning of a real drift.}}}
    \label{fig:drifts_marked}
\end{figure}

Starting the learning process, we sample the data from the stream with a constant chunk size $c$ and monitor the classifier performance using a concept drift detector to detect changes in data distribution. When the drift occurs, we decrease the chunk size to the smaller value $c_d \ll c$,  i.e., $c_d$ is the predefined size of a batch for concept drift. Size of subsequent chunks after drift at given time $t$ are computed using the following equation:
\begin{equation}
    c_{t} =  min(\left\lfloor \alpha c_{t-1} \right\rfloor, c)
\end{equation}
where $\alpha>1$. The chunk size grows continuously with each step to reach the original value $c$ unless the stabilization is detected. Then the chunk size is set to $c$ immediately. 
Let us introduce the \textit{Variance-based Stabilization Detection Method} (VSDM) to detect the predictive performance stabilization. First, we define the fixed-sized sliding window $W$ containing the last $K$ predictive performance metric values obtained for the most recent chunks. We also introduce the stabilization threshold $\epsilon_s$. The stabilization is detected when the following condition is met:
\begin{equation}
    Var(W) < \epsilon_s
\end{equation}

where $Var(W)$ is a variance of scores obtained for the last $K$ chunks. Sample data stream with detected drift and stabilization is presented in Fig.~\ref{fig:drifts_marked}.
The primary assumption of the proposed method is a faster model adaptation caused by the increased number of updates after a \textit{concept drift}. This strategy allows for using the larger chunk sizes when the data is not changing. It also reduces the computational costs of retraining models. Alg.~\ref{caralg} present the whole procedure. 
Our method works with existing models for online learning. For this reason, we argue that the approach proposed in this paper is easier to deploy in practice.

\begin{algorithm}[htbp]
\caption{Chunk-Adaptive Restoration algorithm}
\begin{algorithmic}[1]
\renewcommand{\algorithmicrequire}{\textbf{Input:}}
\renewcommand{\algorithmicensure}{\textbf{Output:}}
\REQUIRE \textit{m} - model \\
\textit{S} - data stream \\
\textit{dd} - drift detector \\
\textit{sd} - stabilization detector \\
\textit{n} - number of chunks \\
\textit{t} - chunk index \\
\textit{c} - base chunk size \\
\textit{$c_d$} - base drift chunk size \\
\textit{$c_t$} - $t$th chunk size\\
\textit{test()} - procedure that tests model with a~chunk and returns the predictive performance metric (ppm) \\
\textit{train()} - procedure that trains model with a~chunk \\
\textit{change\_detected()} - procedure that informs about drift occurrence with the drift detector and the last score \\
\textit{stabilization\_detected()} - procedure that detects stabilization with the stabilization detector and the stabilization window \\
    \FOR {$t = 1$ to $n$}
        \STATE $ppm \leftarrow test(m, S(t))$
        \IF {$stabilization\_detected(sd, ppm)$} 
            \STATE $c_t \leftarrow c$
        \ELSE
            \STATE $c_t \leftarrow min(\left\lfloor \alpha c_{t-1} \right\rfloor, c)$
        \ENDIF
        \IF {$change\_detected(dd, ppm)$}
            \STATE $c_t \leftarrow c_d$
        \ENDIF
        \STATE $train(m, S(t))$
    \ENDFOR
\end{algorithmic}
\label{caralg}
\end{algorithm}

\subsection{Memory and time complexity}

Our method only impacts the size of the chunk. All other factors like the number of features or classifiers in the ensemble are the same as in the basic approach. For this reason, we will focus here only on the impact of chunk size on memory and time complexity. With memory complexity, our method could impact only the size of buffers for storing samples from a stream. When no drift is detected, the standard chunk size is used. This dictates the required size of buffers for storing samples. For this reason, memory complexity for storing samples is $O(c)$.

CAR works the same way as a base method when no drift is detected, and the data stream is stable. Therefore, in this case, the time complexity is the same as in the base method. When drift is detected sizes of subsequent chunks are changed. 
Time complexity depends on model complexity $g(N)$, where $N$ is a number of learning examples provided to model to train on. For simplicity we assume that $g(N)$ represents both ensemble and base model complexity. With this assumptions time complexity of base model (when CAR is not enabled) is: $O(g(c))$.
When CAR is enabled and concept drift is detected chunk size is changed to $c_d$. Each consecutive chunk at time $t$ have size $c_t = \alpha^t c_d$, with $t = 0$ directly after the drift was detected. Chunk size grows until stabilization is detected or current chunk size is restored to original size $c$. For simplicity we skip case when stabilization is detected. With this assumption, we write condition for restoring the original chunk size:

\begin{equation}
    \alpha^{t_s} c_d = c
\end{equation}

Where $t_s$ is time when chunk size is restored to original value. From this equation we obtain $t_s$ directly:

\begin{equation}
    t_s = \log_{\alpha} \frac{c}{c_d}
\end{equation}

The number of operations required by CAR after concept drift was detected is 

\begin{equation}
    \sum_{t=0}^{t_s} g(\alpha^t c_d)
\end{equation}

Using big-O notation:

\begin{equation}
    O(\sum_{t=0}^{t_s} g(\alpha^t c_d)) = O(g(\alpha^{t_s} c_d)) = O(g(\frac{c}{c_d} c_d)) = O(g(c))
\end{equation}

Therefore CAR time complexity depends only on chunk size and computational complexity of used models.

\subsection{Sample Restoration}

Restoration time cannot be directly utilized in this work, as we do not have access to pure streams with separate concepts.
For this reason, we introduce a new Sample Restoration (SR) metric to evaluate the \textit{Chunk-Adaptive Restoration} performance compared to standard methods used for learning models on data streams with concept drift. We assume that there is a sequence of $N$ chunks between two stabilization points. Each element of such a sequence is determined by the chunk size $c_t$ and the achieved model's accuracy $acc_t$. Let us define the index of the minimum accuracy as: 
\begin{equation}
    t_{min} = \underset{t \in [0, N)}{\mathrm{argmin}}\ acc_t
\end{equation}

and the restoration threshold is given by the following formula:
\begin{equation}
    r = p \times \max_{t \in [t_{min}, N)} acc_t
\end{equation}
where $p \in (0, 1)$ is the percentage of the performance that has to be restored, and the multiplier is the maximum accuracy score of our model after the point when it achieved its minimum score.
Finally, we look for the lowest index $t_r$ after which the model exceeds the assumed restoration threshold:
\begin{equation}
    t_r = \inf_{{t \in [t_{min}, N)}} \{ t: acc_{t} \geq r \}
\end{equation}
Sample Restoration is computed as the sum of chunk sizes from the concept drift's beginning to the $t_r$:
\begin{equation}
    SR(p) = \sum_{t=0}^{t_r} c_t
\end{equation}
In general, SR is the number of samples needed to obtain the $p$ percent of the maximum performance achieved on the subsequent task. 

\section{Experiment}

\textit{Chunk-Adaptive Restoration} is a method designed to reduce the number of samples used to restore the model's performance during the \textit{concept drift}. We expect to significantly reduce the \textit{Sample Restoration} for each trained model depending on the chunk size adaptation level. The experimental study was formulated to answer the following research questions:
\begin{itemize}
    \item[RQ1:] How do different chunk sizes impact predictive performance?
    \item[RQ2:] How does the \textit{Chunk-Adaptive Restoration} influence the learning process?
    \item[RQ3:] How many samples can be saved during the restoration phase?
    \item[RQ4:] How do different classifier ensemble models behave with the application of \textit{Chunk-Adaptive Restoration} applied?
    \item[RQ5:] How robouts to noise \textit{Chunk-Adaptive Restoration} is?
\end{itemize}

\subsection{Experiment setup}

\textbf{Data streams.} Experiments were carried out using both synthetic and real datasets. Stream-learn library \cite{ksieniewicz2020stream} was employed to generate the synthetic data containing three types of \textit{concept drift}: abrupt, gradual, and increment, all generated with the recurring or unique concepts. We tested parameters such as chunk sizes and the stream length for each type of \textit{concept drift}. All streams were generated with 5 concept drifts, 2 classes, 20 input features, of which 2 were informative and 2 were redundant. In the case of incremental and gradual drifts concept, sigmoid spacing was set to 5. Apart from the synthetic ones, we employed the Usenet \cite{Katakis2010} and Insects \cite{SouzaChallenges:2020} data streams. Unfortunately, the original Usenet dataset contains a small number of samples, so two selected concepts were repeated to create a recurring-drifted data stream. Each chunk of the Insects data stream was randomly oversampled because of the significant imbalance ratio. Tab.~\ref{tab:stream_variants} contains detailed description of all utilized data streams. 

\begin{table}
    \centering
    \setlength\tabcolsep{6pt}
    \begin{tabular}{llllll}
        \hline
        \bfseries \# & \bfseries Source & \bfseries Drift type & \bfseries \makecell{ Base\\ chunk\\ size $c$ }	& \bfseries \#samples \\ \hline
        1 & stream-learn & abrupt recurring & 500 & 300000 \\
        2 & stream-learn & abrupt recurring & 1000 & 150000 \\
        3 & stream-learn & abrupt recurring & 10000 & 60000 \\
        4 & stream-learn & abrupt recurring & 500 & 250000 \\
        \hline
        5 & stream-learn & abrupt nonrecurring & 500 & 300000 \\
        6 & stream-learn & abrupt nonrecurring & 1000 & 150000 \\
        7 & stream-learn & abrupt nonrecurring & 10000 & 60000 \\
        8 & stream-learn & abrupt nonrecurring & 500 & 250000 \\
        \hline
        9 & stream-learn & gradual recurring & 500 & 300000 \\
        10 & stream-learn & gradual recurring & 1000 & 150000 \\
        11 & stream-learn & gradual recurring & 10000 & 60000 \\
        12 & stream-learn & gradual recurring & 500 & 250000 \\
        \hline
        13 & stream-learn & gradual nonrecurring & 500 & 300000 \\
        14 & stream-learn & gradual nonrecurring & 1000 & 150000 \\
        15 & stream-learn & gradual nonrecurring & 10000 & 60000 \\
        16 & stream-learn & gradual nonrecurring & 500 & 250000 \\
        \hline
        17 & stream-learn & incremental recurring & 500 & 300000 \\
        18 & stream-learn & incremental recurring & 1000 & 150000 \\
        19 & stream-learn & incremental recurring & 10000 & 60000 \\
        20 & stream-learn & incremental recurring & 500 & 250000 \\
        \hline
        21 & stream-learn & incremental nonrecurring & 500 & 300000 \\
        22 & stream-learn & incremental nonrecurring & 1000 & 150000 \\
        23 & stream-learn & incremental nonrecurring & 10000 & 60000 \\
        24 & stream-learn & incremental nonrecurring & 500 & 250000 \\
        \hline
        25 & usenet & abrupt recurring & 1000 & 120000 \\
        26 & insects-abrupt-imbalanced & abrupt nonrecurring & 1000 & 355275 \\
        27 & insects-gradual-imbalanced & gradual nonrecurring & 1000 & 143323 \\
        \hline
    \end{tabular}
    \caption{\fontsize{10pt}{11pt}\selectfont{\itshape{Data streams used for experiments.}}}
    \label{tab:stream_variants}
\end{table}

\noindent \textbf{Drift detector.} The Fast Hoeffding Drift Detection Method 
\cite{Blanco:2015} was employed as a concept drift detector. We used implementation available on the public repository \cite{github}. The size of a window in FHDDM was equal to 1000, and the error probability allowed $\delta$ = 0.000001.

\noindent\textbf{Classifier ensembles.} Three models classifier ensembles dedicated to data stream classification were chosen for comparison: 
\begin{itemize}
    \item Weighted Aging Classifier (WAE) \cite{Wozniak:2013}
    \item Accuracy Weighted Ensemble (AWE) \cite{10.1145/956750.956778},
    \item Streaming Ensemble Algorithm (SEA) \cite{Street:2001},
\end{itemize}

All ensembles contained 10 base classifiers. 

\noindent\textbf{Experimental protocol.} In our experiments, we apply the models mentioned above to selected data streams with concept drift. We measure Sample Restoration. These results are reported as a baseline. Next, we apply Chunk-Adaptive Restoration and repeat experiments to establish the proposed model's influence on the ability to handle concept drift quickly.  As the experiments were conducted with the balanced data, the accuracy was used as the only indicator of the model's performance. As the experimental protocol \textit{Test-Then-Train} was employed \cite{Bifet:2010}.

\noindent \textbf{Statistical analysis.} Because Sample Restoration can be computed for each drift and concept drift can occur multiple times, we report average Sample Restoration for each stream with standard deviation. To assess the statistical significance of the results, we used a one-sided Wilcoxon signed-rank test in a direct comparison between the models with the 95\% confidence level.

\noindent \textbf{Reproducibility.} To enable independent reproduction of our experiments, we provide a github repository with code \footnote{https://github.com/w4k2/chunk-adaptive-restoration}. This repo also contains detailed results of all experiments. Stream-learn \cite{ksieniewicz2020stream} implementation of the ensemble models was utilized with the Gaussian Na\"ive Bayes and CART as base classifiers from sklearn \cite{scikit-learn}. Detailed information about used packages is provided in the yml file with a specification of the conda environment.


\subsection{Impact of chunk size on performance}

\begin{figure}[h]
    \centering
    \includegraphics[width=0.7\textwidth]{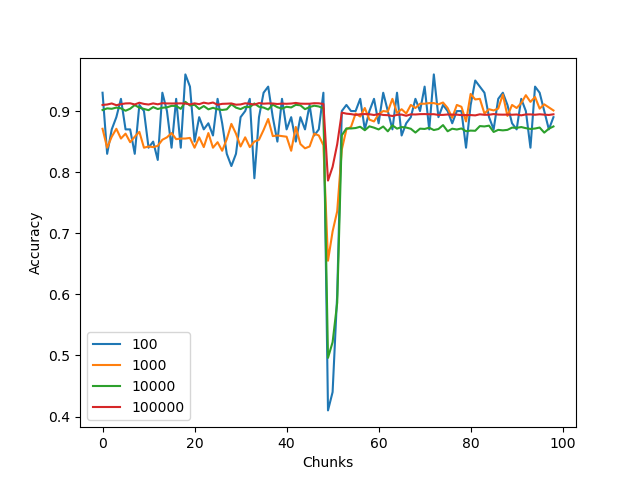}
    \captionsetup{font=small}
    \caption{\fontsize{10pt}{11pt}\selectfont{\itshape{Impact of chunk size on obtained accuracy.}}}
    \label{fig:chunk_size_imapct}
\end{figure}

In our first experiment, we examine the impact of the chunk size on the model performance and general capability for handling data with concept drift. We train the AWE model on a synthetic data stream with different chunk sizes to evaluate these properties. The stream consists of 20 features, 2 classes, and it contains only 1 abrupt drift. Results are presented in Fig.~\ref{fig:chunk_size_imapct}. As expected, chunk size has an impact on the maximal accuracy that the model can achieve. It is especially visible before drift, where models with larger chunks obtain the best accuracy. Also, with larger chunks variance of accuracy is lower. In ensemble-based approaches, a base classifier is trained on a single chunk. A larger chunk means that more data is available to the underlying model. Therefore it allows for the training of a more accurate model.
Interestingly we can see that for all chunk sizes, performance is restored roughly at the same time. Regardless of the chunk size, a similar number of updates is required to bring back the model performance. Please keep in mind that the x-axis in Fig.~\ref{fig:chunk_size_imapct} is the number of chunks. It means that models trained on larger chunks require a larger number of learning examples to restore accuracy.

These results give the rationale behind our method. When drift is detected, we change chunk size to decrease the consumption of learning examples required for restoring accuracy. Next, we gradually increase chunk size to improve the maximum possible performance when the model recovers from drift. It allows for a quick reaction to drift and does not limit the model's maximum performance. In principle, not all models are compatible with changing chunk size. Also, batch size cannot be decreased indefinitely. Minimal chunk size should be determined case by case, dependent on the base learner used in an ensemble or used model in general. Later in our experiments, we use chunk sizes of 500, 1000, and 10000 to obtain a reliable estimate of how our method will perform in different settings.

\subsection{Hyperparameter tuning}

After chunk size was selected, we fine-tuned other hyperparameters, and then we proceeded to further experiments. Firstly set two values manually, based on our observations. First is $\alpha$ (i.e., constant that determines how fast chunk size grows after drift was detected) equal to $1.1$. Second is drift chunk size equal to 30, as it is a typical window length in drift detectors.

Next, we find the best for the stabilization window size and the stabilization threshold. We conduct grid search with windows size values 30, 50, 100, and stabilization thresholds 0.1, 0.01, 0.001, 0.0001. For experiments we use synthetic data streams 1-24 from Tab.~\ref{tab:stream_variants}. Used data streams have different random number generator seeds in this and later experiments. Results were collected for WAE, AWE, SEA ensembles with Na\"ive Bayes base model.
We use Sample Restoration 0.8 as a performance indicator. For each set of parameters, Sample Restoration was averaged over all streams used to obtain one value.
Results are provided in the table \ref{tab:hyperparameters}.

\begin{table}[ht]
    \centering
    \setlength\tabcolsep{6pt}
    \begin{tabular}{cccc}
      \hline
    \multirow{2}{*}{ \bfseries \makecell{stabilization\\ thresholds}} & \multicolumn{3}{c}{ \bfseries drift chunk size} \\
    & \bfseries 30 & \bfseries 50 & \bfseries 100 \\
      \hline
      0.1 & 59210.11 & 59210.11 & 59210.11 \\
      0.01 & 58489.47 & 58675.99 & 58709.98 \\
      0.001 & 55328.20 & 55363.95 & 57669.70 \\
      0.0001 & 52846.04 & 55962.58 & 62398.56 \\
      \hline
    \end{tabular}
    \caption{\fontsize{10pt}{11pt}\selectfont{\itshape{Sample Restoration 0.8 for various hyperparameter setting. Lower is better.}}}
    \label{tab:hyperparameters}
\end{table}

From provided data, we can conclude that the smaller the drift chunk size, the lower the SR is. This observation is in line with intuition about our method. Smaller drift chunk size provides a larger benefit during drift compared to normal chunk size. The same dependency can be observed for the stabilization threshold. Intuitively, a lower threshold means that stabilization is harder to reach. We argue that this can be beneficial in some cases when working with gradual or incremental drift. In this scenario, if stabilization is reached too fast, then chunk size is immediately brought back to the standard size, and there is no benefit from a smaller chunk size at all. Lowering the stabilization threshold could help in these cases.
In later experiments, we use the stabilization window size equal to 30 and the variance stabilization threshold equal to 0.0001.

\subsection{Impact on concept drift handling capability}

\begin{figure}[h]
  \centering
  \begin{minipage}[b]{0.48\textwidth}
    \includegraphics[width=\textwidth]{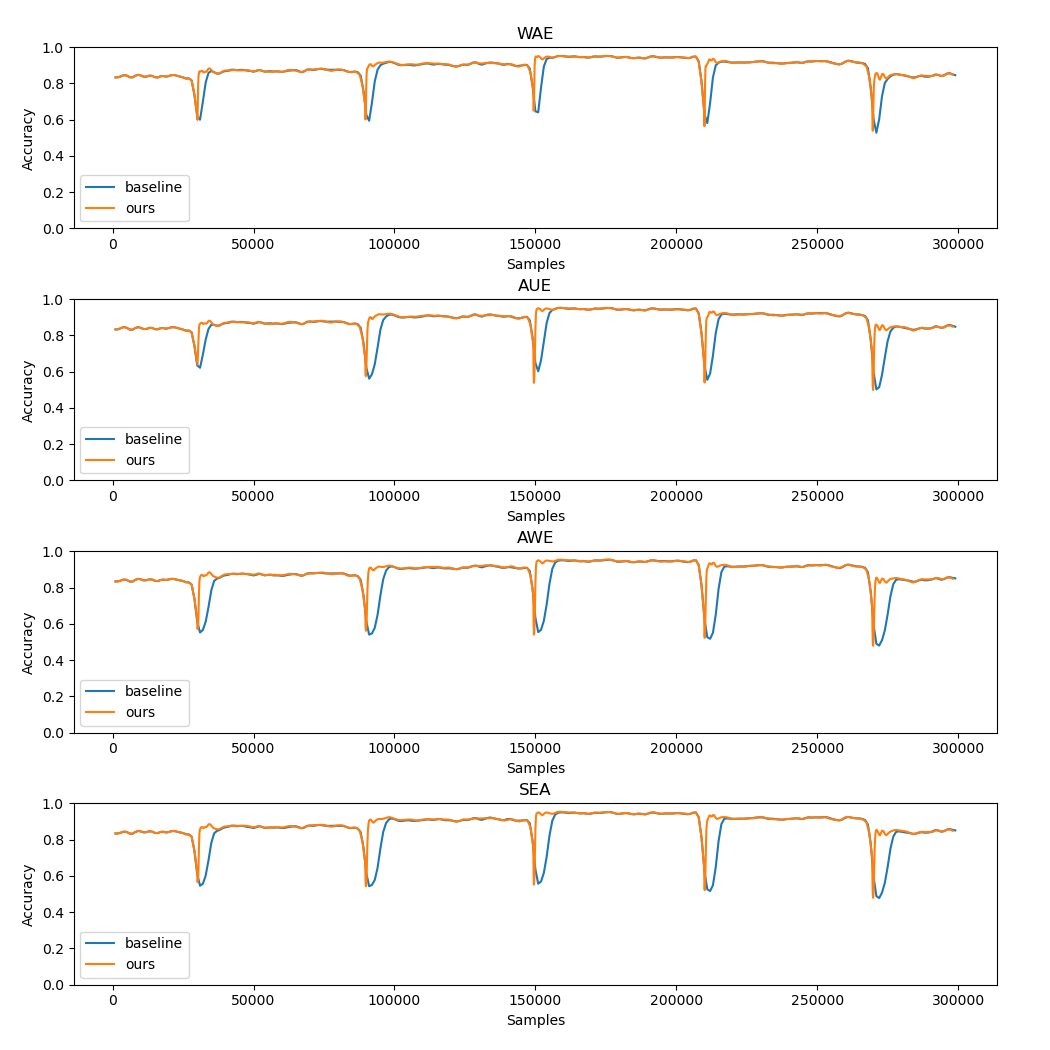}
    \captionsetup{font=small}
    \caption{\fontsize{10pt}{11pt}\selectfont{\itshape{Accuracy for stream-learn data stream (1).}}}
    \label{fig:strlearn}
  \end{minipage}
  \hfill
  \begin{minipage}[b]{0.48\textwidth}
    \includegraphics[width=\textwidth]{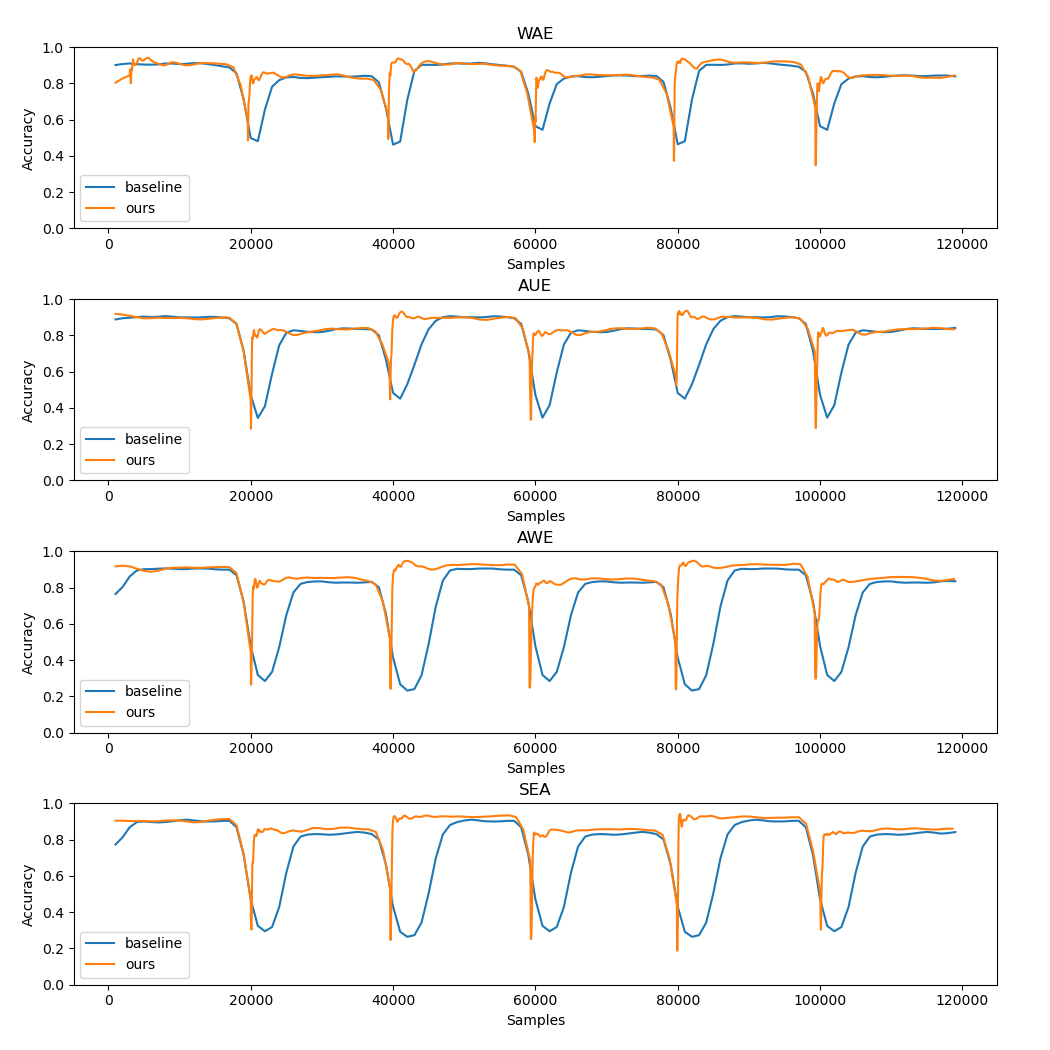}
    \captionsetup{font=small}
    \caption{\fontsize{10pt}{11pt}\selectfont{\itshape{Accuracy for Usenet dataset (25).}}}
    \label{fig:usenet}
  \end{minipage}
\end{figure}

\begin{figure}[!htbp]
  \centering
  \begin{minipage}[b]{0.48\textwidth}
    \includegraphics[width=\textwidth]{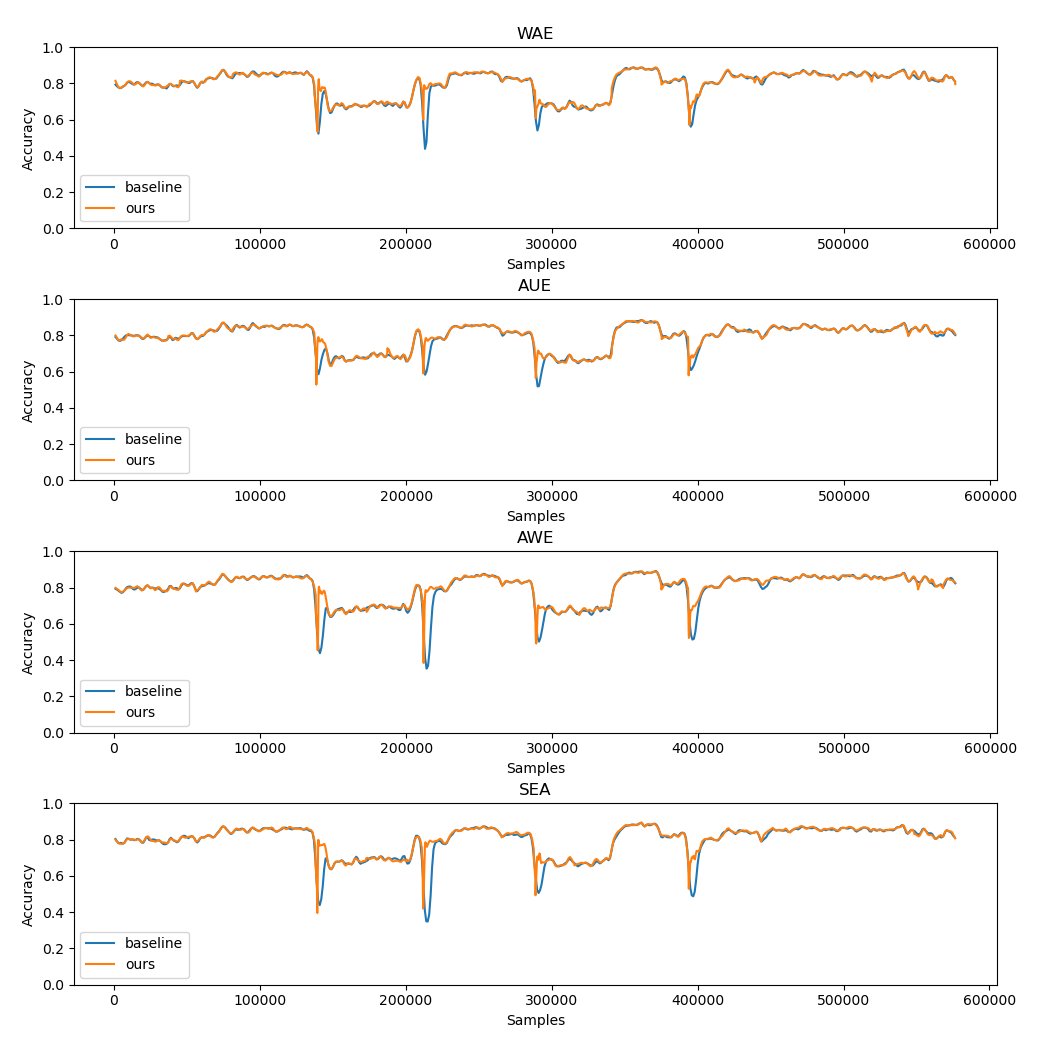}
    \captionsetup{font=small}
    \caption{\fontsize{10pt}{11pt}\selectfont{\itshape{Accuracy for abrupt Insects dataset (26).}}}
    \label{fig:insects_abrupt}
  \end{minipage}
  \hfill
  \begin{minipage}[b]{0.48\textwidth}
    \includegraphics[width=\textwidth]{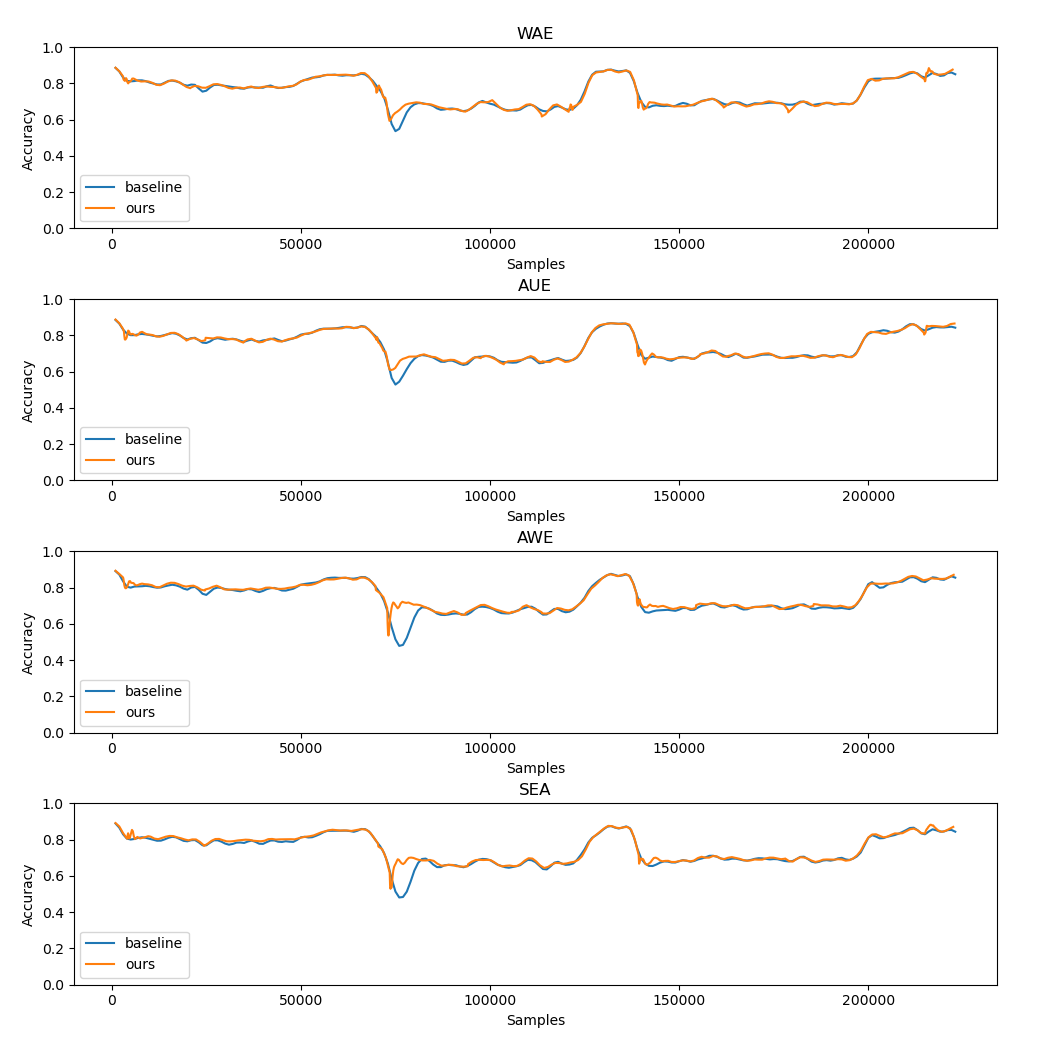}
    \captionsetup{font=small}
    \caption{\fontsize{10pt}{11pt}\selectfont{\itshape{Accuracy for gradual Insects dataset (27).}}}
    \label{fig:insects_gradual}
  \end{minipage}
\end{figure}

In this part of the experiments, we compare the performance of the proposed method to baseline. Results were collected following the experimental protocol described in the previous sections. To save space, we do not provide results for all models and streams. Instead, we plot accuracy achieved by models on selected data streams. These results are presented in Fig.~\ref{fig:strlearn}, ~\ref{fig:usenet}, ~\ref{fig:insects_abrupt}, and ~\ref{fig:insects_gradual}. All learning curves were smoothed using a 1D Gaussian filter with $\sigma=1$. 

From provided plots, we can deduce that the largest gains from employing the CAR method can be observed for an abrupt data stream. In streams with gradual and incremental drifts, there are fewer or none sudden drop of accuracy that the model can quickly react to. For this reason, the CAR method does not provide a large benefit with this kind of concept drifts. During a more detailed analysis of obtained results, we observed that the stabilization for gradual and incremental drifts is hard to detect. Many false positives usually cause an early return to the original chunk size, influencing the performance achieved on those two types of drifts. FHDDM caused another problem regarding the early detection of the gradual and incremental concept drifts. Usually, this is a desired feature. In our method, early drift detection initiates the chunk size change when two data concepts are still overlapping during stream processing. As the transition between two concepts takes much time, when one concept starts to dominate, the chunk size could be restored to its original value too early, affecting the achieved results.

We also observe larger gains from applying CAR on streams with bigger chunk size. To illustrate please compare results from Fig.~\ref{fig:strlearn} to Fig.~\ref{fig:usenet}. One possible explanation behind this trend is that gains obtained from employing CAR are proportional to the difference in size between the base and drift chunk size. In our experiments, drift chunk size was equal to 30 for all streams and models. This explanation is also in line with the results of hyperparameter experiments provided in Tab.~\ref{tab:hyperparameters}.

We conclude this section by providing a statistical analysis of our results. Tab.~\ref{tab:stats} shows the results of the Wilcoxon test for Na\"ive Bayes and CART base models. We state meaningful differences in the Sample Restoration between the baseline and the CAR method for all models.


\begin{table}[ht]
    \centering
    \setlength\tabcolsep{6pt}
    \begin{tabular}{ccccccc}
      \hline
      \multicolumn{7}{c}{ \bfseries Na\"ive Bayes } \\
      \hline
      \multirow{2}{*}{ \bfseries \makecell{model\\ name}} & \multicolumn{2}{c}{ \bfseries SR(0.9)} & \multicolumn{2}{c}{ \bfseries SR(0.8)} & \multicolumn{2}{c}{ \bfseries SR(0.7)} \\
      & \bfseries Statistic & \bfseries p-value & \bfseries Statistic & \bfseries p-value & \bfseries Statistic & \bfseries p-value \\ \hline
      WAE & 40.0 & 0.0006 & 30.0 & 0.0002 & 45.0 & 0.0009 \\ 
      AWE & 22.0 & 9.675e-05 & 26.0 & 0.0001 & 36.0 & 0.0004 \\ 
      SEA & 0.0 & 1.821e-05 & 23.0 & 0.0001 & 1.0 & 1.389e-05 \\ 
      \hline
      \multicolumn{7}{c}{ \bfseries Cart } \\
      \hline
      \multirow{2}{*}{ \bfseries \makecell{model\\ name}} & \multicolumn{2}{c}{ \bfseries SR(0.9)} & \multicolumn{2}{c}{ \bfseries SR(0.8)} & \multicolumn{2}{c}{ \bfseries SR(0.7)} \\
      & \bfseries Statistic & \bfseries p-value & \bfseries Statistic & \bfseries p-value & \bfseries Statistic & \bfseries p-value \\ \hline
          WAE & 14.0 & 6.450e-05 & 54.0 & 0.003 & 55.0 & 0.003 \\ 
          AWE & 0.0 & 1.229e-05 & 6.0 & 2.543e-05 & 21.0 & 0.0001 \\ 
          SEA & 23.0 & 0.0001 & 43.0 & 0.001 & 42.0 & 0.001 \\ 
      \hline
    \end{tabular}
    \caption{\fontsize{10pt}{11pt}\selectfont{\itshape{Wilcoxon test results}}}
    \label{tab:stats}
\end{table}

\subsection{Impact of noise on the CAR effectiveness}

\begin{figure}[h]
    \centering
    \includegraphics[width=0.8\textwidth]{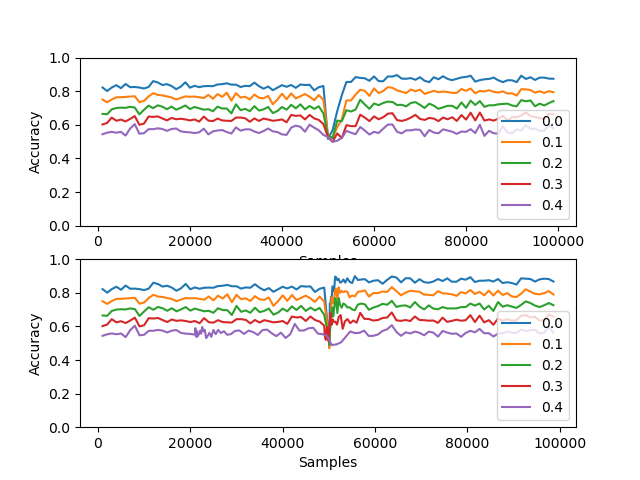}
    \captionsetup{font=small}
    \caption{\fontsize{10pt}{11pt}\selectfont{\itshape{Impact of noise in labels on proposed method effectiveness. (Upper) baseline accuracy for synthetic data stream with different noise level added to labels. (Lower) CAR accuracy for the same synthetic data stream. In case of Noise levels 0.2, 0.3, and 0.4 stabilization detector was turned off.}}}
    \label{fig:noise_imapct}
\end{figure}

Real-world data often contain noise in labeling. For this reason, we evaluate if the proposed method can be used for data with varying amounts of noise in labels. We generate a synthetic data stream with two classes, base chunk size 1000, drift chunk size 100, and single, abrupt concept drift. We randomly select a predefined fraction of samples in each chunk and flip labels for selected learning examples. Next, we measure the accuracy of the AUE model with Gaussian Na\"ive Bayes base model on a generated dataset with noise levels 0, 0.1, 0.2, 0.3, and 0.4. Results are presented in Fig.~\ref{fig:noise_imapct}. We note for low levels of noise i.e., up to 0.3, restoration time is shorter. 
With a larger amount of noise, there is no sudden drop in accuracy. Therefore CAR has no impact on the speed of reaction to drift. 

It should be noted that results for CAR with noise levels 0.2, 0.3, and 0.4 were generated with the stabilization detector turned off. With a higher amount of noise, stabilization was detected very fast. Therefore chunk size was quickly set to base value. In this case, there was no benefit of applying CAR. This indicates that the stabilization method should be refined to handle noisy data well.

\subsection{Lessons learned}
Firstly we evaluated the impact of chunk size on the process of learning in the data stream with single concept drift. We learn that models with larger chunk size can obtain larger maximum accuracy, but the required number of updates to restore accuracy is similar regardless of chunk size (RQ1 answered).
The main goal of introducing the \textit{Chunk-Adaptive Restoration} was to prove its advantages in controlling the number of samples during the restoration period while dealing with abrupt \textit{concept drift}. The statistical tests have shown a significant benefit of employing it in different stream learning scenarios (RQ2 answered). 
The highest gains of employing the method were observed when the large original chunk size was used. With a bigger chunk size, there are fewer model updates, resulting in a delay of reaction to concept drift.

The number of samples that can be saved depends on the drift type and the original chunk size. When dealing with abrupt drift, the sample restoration time can be around 50\% better than the baseline (RQ3 answered).
We noticed that for each of the analyzed classifier ensemble methods, CAR minimized restoration time and achieved better average predictive performance. It is worth noting that the simpler the algorithm, the greater the profit from using CAR. The most considerable profit was observed for SEA and AWE, while in the case of WAE, sometimes the native version outperformed CAR for the Average Sample Restoration metric (RQ4 answered).
When a small amount of noise is present in labels, CAR can still be useful, however in some cases stabilization detector should not be used. With a larger amount of noise, there is no gain from using the proposed method (RQ5 answered).


\section{Conclusions}

The work focused on the \textit{Chunk-Adaptive Restoration} framework, which is dedicated to chunk-based data stream classifiers enabling better recovery from concept drifts. To achieve this goal, we proposed new methods for stabilization detection and chunk size adaptation. Their usefulness was evaluated based on computer experiments conducted on the real and synthetic data streams. Obtained results show a significant difference between the predictive performance of the baseline models and models employing CAR. \textit{Chunk-Adaptive Restoration} is strongly recommended for abrupt concept drift scenarios because it significantly can reduce model downtime. The performance gain is not visible for other types of concept drift, but it still achieves acceptable results.
The future works may focus on:
\begin{itemize}
    \item Improving the \textit{Chunk-Adaptive Restoration} behavior for gradual and incremental concept drifts.
    \item Adapting the \textit{Chunk-Adaptive Restoration} to the case of limited access to labels using a semi-supervised and active learning approach.
    \item Proposing a more flexible method of changing data chunk size, e.g.,  based on the model stability assessment.
    \item Adapting the proposed method to imbalanced data stream classification task, where changing the data chunk size may be correlated with the intensity of data preprocessing (e.g., the intensity of data oversampling).
    \item Improve stabilization method to better handle data streams with noise.
\end{itemize}

\section*{Acknowledgement}
This work is supported by the CEUS-UNISONO programme, which has received funding from the National Science Centre, Poland under grant agreement No. 2020/02/Y/ST6/00037.

\end{document}